\documentclass{article} \newcommand{\debug}[1]{} 
\usepackage{aaai}
\usepackage{logic,latexsym,xspace,eufrak}
\usepackage{amsmath,amsthm}
\usepackage[dvips]{epsfig}
\usepackage{afterpage,wrapfig}

\providecommand{\debug}[1]{}
\providecommand{\final}[1]{#1}

\author{Steffen H\"olldobler and Hans--Peter St\"orr \\
  Artificial Intelligence Institute \\ 
  Department of Computer Science \\ 
  Dresden University of Technology \\
  \{sh,hans-peter\}@inf.tu-dresden.de\\ }
\date{}
\title{BDD--based Reasoning in the Fluent Calculus --- First Results}

\catcode`°=\active
\def°{{\ensuremath{\kringel}}}
\newcommand{\kringel}{\,\circ\,}
\newcommand{\Emp}{\emptyset}

\newcommand{\ignore}[1]{}
\debug{\usepackage{showlabels}}

\debug{
  \newcommand{\todo}[1]{{\scriptsize\sc ToDo: #1}}
} \providecommand{\todo}[1]{}

\newcommand{\ZZ}{\mbox{\bf Z}}

\newcommand{\Goal}{{\ensuremath{\Phi_G}}}

\newcommand{\Not}{\neg}

\newcommand{\zv}{\vec{z}}
\newcommand{\zpv}{\vec{z}\,'}

\newcommand{\Set}[1]{{\ensuremath{\left\{#1\right\}}}\xspace} 

\newcommand{\Subst}[1]{{\ensuremath{\{#1\}}}\xspace} 
\newcommand{\True}{{\ensuremath{\top}}\xspace}
\newcommand{\False}{{\ensuremath{\bot}}\xspace}

\renewcommand{\Exists}[1]{(\exists #1)\ }

\newcommand{\Srm}{\mbox{\sc room\/}} 
\newcommand{\Sbl}{\mbox{\sc ball\/}}
\newcommand{\Sgr}{\mbox{\sc gripper\/}}

\newcommand{\ATR}{\mbox{\em at-robby}} 
\newcommand{\AT}{\mbox{\em at}}
\newcommand{\FREE}{\mbox{\em free}}
\newcommand{\CARRY}{\mbox{\em carry}}

\newcommand{\MOVE}{\mbox{\em move}} 
\newcommand{\PICK}{\mbox{\em pick}}
\newcommand{\DROP}{\mbox{\em drop}}
\newcommand{\NOOP}{\mbox{\em noop}}

\newcommand{\RA}{A} 
\newcommand{\RB}{B}

\newcommand{\GA}{G_1} 
\newcommand{\GB}{G_2}

\newcommand{\BONE}{B_1}

\newcommand{\Bn}{B_n}

\newcommand{\BDDPLAN}{\textsc{BDDplan}} 

\theoremstyle{plain}
\newtheorem{theorem}{Theorem}
\newtheorem{lemma}[theorem]{Lemma}

\newcommand{\comment}[1]{
    \marginpar[\raggedleft{}\footnotesize{\hspace{0pt}\em #1}]%
    {\raggedright{}\footnotesize{\hspace{0pt}\em #1}}}

\newcommand{\FF}{\mbox{\bf F}}

\newcommand{\TT}{\mbox{\bf T}}

\newcommand{\CalB}{{\cal B}}

\newcommand{\CalF}{{\cal F}}

\newcommand{\CalZ}{{\cal Z}}

\newcommand{\DO}{\mbox{\em do\/}}

\newcommand{\HLD}{\mbox{\em holds\/}}

\newcommand{\STT}{\mbox{\em state\/}}

\newcommand{\Sac}{\mbox{\sc action\/}}
\newcommand{\Ssi}{\mbox{\sc sit\/}}

\newcommand{\Sfl}{\mbox{\sc fluent\/}}
\newcommand{\Sst}{\mbox{\sc state\/}}

\renewcommand{\comment}[1]{
    \marginpar[\raggedleft{}\tiny\sf\parbox{1cm}{#1}]
    {\raggedright{}\footnotesize{\mbox{}\hspace{-2mm}\tiny\sf\parbox{0.8cm}{#1}}}}
\final{\renewcommand{\comment}[1]{}}

\begin{document}

\nocopyright

  \maketitle

\begin{abstract}
  \noindent The paper reports on first preliminary results and
  insights gained in a project aiming at implementing the fluent
  calculus using methods and techniques based on binary decision
  diagrams. After reporting on an initial experiment showing
  promising results we discuss our findings concerning various
  techniques and heuristics used to speed up the reasoning process.
  \vspace{3mm}

  \noindent
  Keywords: Reasoning about Actions, Fluent Calculus, Planning, Binary Decision Diagrams.\end{abstract}

\section{Introduction}
\label{s:intro}
In recent years we have seen highly advanced and novel
implementations\comment{formulas oder formulae?}
of propositional calculi and systems like, for example, {\sc GSat\/}
and its variants \cite{selman:etal:92}, {\sc Smodels}
\cite{niemelae:simons:97} or {\sc DLV} \cite{eiter:etal:98}, to
mention just a few. The implementations were applied to many
interesting fields in Intellectics like, for example, planning or
non--monotonic reasoning. On the other hand, few results are
reported so far on applying another propositional method in these
fields, viz., model checking using binary decision diagrams (BDDs),
with Cimatti et. al \shortcite{cimatti:etal:97,cimatti:etal:98,cimatti:rovieri:99} and \cite{edelkamp:reffel:99} being an exception. This
comes to a surprise because model checking using binary decision
diagrams has significantly improved the performance of algorithms and
enabled the solution of new classes of problems in related areas like
formal verification and logic synthesis (see e.g.\
\cite{burch:etal:92,burch:etal:94}). Can we adopt the technology
developed for model checking of finite state machines using binary
decision diagrams for the solution of planning problems and, more
generally, problems occurring in reasoning about situations, actions
and causality\/? Can we enrich these techniques by exploiting the
experiences made in the state of the art implementations of
propositional logic calculi and systems mentioned at the beginning of
this paragraph\/?

In order to answer these and related questions a sound and complete mapping from
(a fragment of) the fluent calculus \cite{hs:90,thielscher:98} to propositional
logic is specified in \cite{hoelldobler:stoerr:99a} such that the entailment
problem in the fluent calculus can be solved by finding models for the
corresponding propositional logic formula. The propositional logic formulae are
represented by reduced and ordered binary decision diagrams and techniques from
model checking are applied to search for models.  Our mapping relies on three
properties of the considered fragment of the fluent calculus:
\begin{itemize}
\item The set of states is characterized by a finite set of propositional
      fluents, i.e., a set of propositional variables, which can take
      values out of $\{\top,\bot\}$.

\item The actions are deterministic and their preconditions as well as effects
      depend only on the state they are executed in.

\item The goal of the planning problem is a property which depends solely on
      the reached state.
\end{itemize}

Here we report on initial results, findings and insights gained with
the BDD--based implementation of the fluent calculus. After briefly
discussing the fluent calculus and the implementation using an example
from the so--called {\sc Gripper}--class, we concentrate on two
heuristics and techniques which can be applied to speed up the solution of the
planning problem. In particular, we discuss some results on variable ordering
and partitioning of the transition relation.

For the convenience of the reader we give a very brief review of some basic
notions of the fluent calculus and the concept of BDDs, but for a detailed
introduction into the matter we refer to \cite{thielscher:98,hoelldobler:stoerr:99a} and
\cite{bryant:86} as references respectively.

\section{Gripper Planning Problems}

In a contest held at AIPS98, planners had to solve various problems,
among which were the problems of the so--called {\sc Gripper} class: 
\begin{quote}
  {\em A robot equipped with two grippers $\GA$ and $\GB$ can move
  between two rooms $\RA$ and $\RB$. Initially the robot is in room
  $\RA$ together with a number of balls $\BONE,\ \ldots, \Bn$. The
  task is to transport these balls into room $\RB$.\/} 
\end{quote}

We will specify {\sc Gripper} class problems in the fluent calculus in a moment.
Before doing so, however, some notational conventions are helpful. Words
starting with an upper letter denote constants, whereas words starting with a
lower letter denote predicate symbols, non--nullary function symbols and
variables. Additionally we assume that each variable $a$ denotes an action, $s$
a situation, $f$ a fluent and $z$ a state (i.e. are variables of the
corresponding sorts $\Sac,\Ssi,\Sfl,\Sst$ in the fluent calculus, see
\cite{thielscher:98,hoelldobler:stoerr:99a}). All symbols may be indexed.

In the fluent calculus situations are denoted by $S_0$ standing for the initial
situation, and by use of the function $\DO(a,s)$, denoting the situation after
execution of an action $a$ in an situation $s$. States are denoted by combining
the fluents, which hold in the state, with the associative and commutative
binary operation symbol~$\circ$, effectively representing multisets of fluents.
In the case of propositional fluents, as considered in this paper, the states
contain each fluent at most once. $\Emp$~represents the empty state. Thus, $°$
fulfills the properties:\footnote{Free variables in formulae are assumed
  universally quantified, unless otherwise stated.}
\begin{equation} \label{e:ac1} \tag{AC1}
  \begin{array}{rcl}
  (z_1°z_2)°z_3 &=& z_1°(z_2°z_3) \\
  z_1°z_2 &=& z_2°z_1 \\
  z°\Emp &=& z
  \end{array}
\end{equation}
A normal form\comment{``Normalform'' suggeriert Eindeutigkeit.
  Besserer Begriff?} to write ground state terms are the so--called
{\em constructor state terms\/} of the form $\emptyset \circ f_1\circ \ldots
\circ f_n$, $n \geq 0$ where the $f_i\!$'s are pairwise distinct.

$\STT(s)$ denotes the state holding in a
situation $s$.  We also make frequently use of the abbreviation
\begin{displaymath}  
  \HLD(f,s) \equiv \Exists{z} \STT(s)=f\circ z \enspace .
\end{displaymath}

The initial state of a reasoning problem in the fluent calculus is specified
by an axiom of the form
\begin{equation} \label{e:szero}
\CalF_{S_0} = \{\STT(S_0) = t\},
\end{equation}
relating the initial situation $S_0$ to a state $t$ represented as an constructor
state term. If an equation like~(\ref{e:szero}) is given, then $\Phi_I(z)$
denotes the equation $z = t$. Turning to the example, the initial state of a
{\sc Gripper} class problem is specified by 
\begin{multline*}
\CalF_{S_0} = \{ \STT(S_0) = \Emp°\AT(\BONE,\RA)°\ldots°\AT(\Bn,\RA) \\ 
\circ\,  \FREE(\GA) \circ \FREE(\GB) \circ \ATR(\RA) \},
\end{multline*}
where $n$ is instantiated to some number.

The fluent $\AT(b,r)$ states
that ball $b$ is at room $r$, $\FREE(g)$ states that gripper $g$ is
free and $\ATR(r)$ states that the robot is at room~$r$. \footnote{Formally,
  $b$, $r$, $g$ denote variables of new sorts $\Sbl$, $\Srm$ and $\Sgr$.}

There are three actions in the {\sc Gripper} class:
\begin{itemize}
\item the robot may {\em move\/} from one room to the other. 
\item the robot may {\em pick\/} up a ball if it is in the same room
      as the ball and one of its grippers is empty.
\item the robot may {\em drop\/} a ball if it is carrying one.
\end{itemize}
These actions are specified by means of state update axioms, which relate a
state $\STT(s)$ and the state $\STT(\DO(a,s))$ after executing an action $a$.
The general form for state update axioms is as follows:
\begin{displaymath} 
    \Delta(s) \IMPL \STT(\DO(a,s)) \circ \vartheta^- = \STT(s) \circ
    \vartheta^+ \enspace ,
\end{displaymath}
where $\vartheta^+$ are positive Effects of $a$, i.e. the fluents
which did not hold before and will hold after executing the action,
$\vartheta^-$ negative Effects and $\Delta(s)$ is the condition
under which the action has exactly these Effects.
For technical reasons we include an action $\NOOP$ which leaves the state
unchanged.

\begin{displaymath}
\begin{array}{r@{}c@{}l@{}l@{}l}
\CalF_{su} &= \{ & \HLD(\!\ATR(r_1),s)\! \AND\! \Not\HLD(\!\ATR(r_2),s) \\
& & \IMPL\ \STT(\DO(\MOVE(r_1,r_2),s))°\ATR(r) \\
& & \hspace{7mm} =\ \STT(s)°\ATR(r_2) & ,\\[2mm]
& & \HLD(\AT(b,r),s) \AND \HLD(\ATR(r),s) \\
& & \AND\ \HLD(\FREE(g),s) \AND \Not\HLD(\CARRY(b,g),s) \\
& & \IMPL\ \STT(\DO(\PICK(b,r,g),s))°\AT(b,r)°\FREE(g) \\
& & \hspace{7mm} =\ \STT(s)°\CARRY(b,g) & ,\\[2mm]
& & \HLD(\CARRY(b,g),s) \AND \HLD(\ATR(r),s)\\
& & \AND\ \Not\HLD(\AT(b,r),s) \AND \Not\HLD(\FREE(g),s)\\
& & \IMPL \STT(\DO(\DROP(b,r,g),s))°\CARRY(b,g) \\
& & \hspace{7mm} =\ \STT(s)°\AT(b,r)°\FREE(g) & , \\[2mm]
& & \STT(\DO(\NOOP,s)) = \STT(s) & \}
\end{array}
\end{displaymath}
The states, which may occur in a planning problem, are constrained the following
axiom, which states, that fluents cannot occur twice in a state:
\begin{equation*}   \label{eq:fms}
  \CalF_{ms} = \{ (\forall s,\ z)\ \neg (\exists g)\ \STT(s) = g \circ g \circ z
  \} \enspace .
\end{equation*}
The complete fluent calculus axiomatization $\CalF$ of a planning problem
consists of the axioms mentioned above, as well as some additional axioms, whose
discussion is out of the scope of this paper, namely a set $\CalF_{mset}$ of
equational axioms containing \eqref{e:ac1}, which define the semantics of
$\circ$, as well as a set of unique name assumptions $\CalF_{un}$ for the sort
$\Sfl$. Please refer to the given literature for a detailed discussion.
\begin{equation*} \label{eq:defcalf}
  \CalF = \CalF_{un} \cup \CalF_{mset} \cup \CalF_{S_0} \cup \CalF_{ms} \cup
  \CalF_{su} \enspace.
\end{equation*}
 
 Reasoning problems themselves are specified as entailment problems in the
fluent calculus. For the {\sc Gripper} class we obtain the entailment problem
\begin{displaymath}
\CalF \models (\exists s)\ \HLD(\AT(\BONE,\RB),s) \wedge \ldots \wedge
\HLD(\AT(\Bn,\RB),s).
\end{displaymath}

In general, reasoning about planning problems in the fluent calculus amounts to
solving an entailment problem of the form
\begin{displaymath}
\CalF \models \Exists{z} [(\exists s)\ \STT(s) = z] \wedge \Phi_G(z),
\end{displaymath}
where $\Phi_G(z)$ is a goal formula with $z$ as the only free variable.
Such problems have a solution if we find a substitution $\sigma$ for $z$ 
such that
\begin{gather}
  \label{e:reachable}
  \CalF\models [\Exists{s} \STT(s) = z \sigma]\\
  \intertext{and}
  \label{e:goal}
  \CalF \models \Goal(z\sigma) \enspace .
\end{gather}
It is sufficient to restrict our search to substitutions $\sigma$ which actually
denote states of our reasoning problem, i.e., substitutions which contain solely bindings of variables of sort
$\Sst$ to constructor state 
terms. Such substitutions are called {\em constructor state
substitutions\/}. In the sequel, $\sigma$ will always denote a constructor
state substitution.

The main idea of our algorithm presented in \cite{hoelldobler:stoerr:99a,hoelldobler:stoerr:00a} is to calculate
successively a sequence $(\CalZ_i\mid i \geq 0)$ of sets of solutions to \eqref{e:reachable} which
correspond to the sets of states reached after executing $0,\ 1,\ 2,\ \dots$ actions
starting in the initial state, until a state is found which is a goal state,
i.e.\ the corresponding substitution fulfills \eqref{e:goal}, or, if no new states are reached, in which case
there is no plan. The implementation of this algorithm is done by representing these sets as well as the
relation between them by means of binary decision diagrams (BDDs).

\subsection{Binary Decision Diagrams}

The idea of BDDs is similar to decision trees: a Boolean function is
represented as a rooted acyclic directed graph. The difference to decision
trees is that there is a fixed order of the occurrences of variables in each
branch of the diagram, and that isomorphic substructures of the diagram are
represented only once.\footnote{Thus, the BDD is \emph{ordered} and
  \emph{reduced}, also called ROBDD. These properties are so useful that they are required in
  almost all BDD applications, so many authors include these properties into
  the definition of BDDs.} This can lead to exponential savings in
space in comparison to representations like decision trees or disjunctive or
conjunctive normal form.

We will introduce BDDs via an example.  A formal treatment of BDDs is out of the
scope of this paper and we refer the interested reader to the literature (see
e.g.\ \cite{bryant:86,clarke:etal:94,burch:etal:92}).

\begin{wrapfigure}{r}{1.9cm}
  \mbox{}\hspace{-0.4cm}\epsfig{file=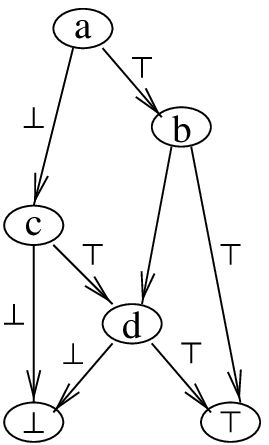,width=2.2cm}\vspace{0cm}\mbox{}
\end{wrapfigure}
Consider the propositional logic formula $(a\wedge
b)\vee(c\wedge d)$. Using the variable ordering $a<b<c<d$ a BDD
representation of this formula is given in the figure to the right. For a given valuation of the propositional variables $a,\ b,\ c$ and $d$ the
value of the Boolean function represented by the BDD is obtained by traversing
the diagram starting from the root and taking at each node the edge labeled
with the value of the variable occurring in the node.

The label of the
terminal node defines the value of the function under the current valuation.
For example $\left\langle a\mapsto \bot,\ b \mapsto \bot,\ c\mapsto \top,
  d\mapsto \bot \right\rangle$ leads to a node labeled $\bot$, i.e., the value
of the formula is $\bot$ wrt this valuation.

Bryant has shown in \cite{bryant:86} that, given a fixed variable order, every
Boolean function is represented by exactly one BDD. Moreover, propositional
satisfiability, validity and equivalence problems are decidable over BDDs in
linear or constant time. Of course, the complexity of the mentioned problems
does not go away: the effort has been moved to the construction of the BDDs.
But as Bryant has shown as well, there are efficient algorithms for logical
operations, substitutions, restrictions etc.\ on BDDs, whose cost is in most
cases proportional to the size of its operands.  BDDs may be used as a theorem
prover, i.e., by construction of a BDD corresponding to a logical formula, and
check the BDD for interesting properties, but more often they are used as an
implementation tool for algorithms which are semantically based on Boolean
functions or, equivalently, propositional formulae, or, via the characteristic
functions, sets. In the implementation these formulae or sets are always
represented as BDDs.  The use of BDDs in this paper follows this spirit.

\section{The Algorithm}\label{s:alg}

The algorithm for solving entailment problems in the fluent calculus follows
in spirit the algorithm to find reachable states in finite state systems as presented e.g. in
\cite{burch:etal:94}. As mentioned, the aim is to find the sets $\CalZ_i$ of
solutions for \eqref{e:reachable} representing states which can be reached
from the initial state after the execution of $i$ actions. The first crucial
question to tackle is how to represent these sets using BDDs.

Each solution to \eqref{e:reachable} is a constructor state substitution
$\Subst{z/t}$ with a term $t$ of the form $\emptyset \circ f_1\circ \ldots
\circ f_n$, where the $f_i\!$'s are pairwise distinct. On first glance it
seems impossible to represent substitutions by finite BDDs because
there are infinitely many terms. Fortunately, however, if there are only
finitely many fluents then there are also only finitely many terms $t$ such
that $\{z / t\}$ satisfies \eqref{e:reachable} due to
$\CalF_{ms}$. Furthermore, because $\circ$ is an AC1--symbol in the fluent
calculus we do not have to distinguish between terms which are equivalent
under the AC1 equational theory. In other words, a term $t$ occurring in the
codomain of a constructor state substitution is uniquely characterized by the
set of fluents occurring in~$t$.

This observation opens a possibility for encoding sets of solutions for the
entailment problem in the fluent calculus into a BDD: for each of the finitely
many fluents $f$ which may occur in the binding for a variable $z$ in a constructor state
substitution we introduce a propositional variable $z_f$. A constructor state
substitution $\sigma=\Subst{z/t}$ is represented by a valuation
$\CalB_S(\sigma)$ for these variables such that $z_f$ is mapped to $\True$ by
$\CalB_S(\sigma)$ iff $f$ occurs in $t$. \footnote{A substitution containing
  more than one binding is represented similarly: for each variable in the
  domain of the substitution we introduce a separate set of propositional
  variables which encodes the binding of that variable.}
Hence, a set $S$ of constructor state
substitutions is represented by a set of valuations. The set of valuations
itself is represented by a propositional formula $Z$ such that the
set of models for $Z$ is the set of valuations. Finally, $Z$ is represented
by a BDD. For example, if the alphabet underlying the fluent calculus contains 
precisely the fluent symbols $a,\ b$ and $c$, then a substitution
$\sigma=\Subst{z/a°c}$ is represented by a valuation as follows:
\begin{displaymath}
  \begin{array}{rclcccl}
    \sigma&=& \{z/ & a° &  & c & \}\\
    \CalB_S(\sigma)&=&\{& z_a\mapsto\True, & z_b\mapsto\False,&z_c\mapsto\True&\}
    \enspace ,
  \end{array}
\end{displaymath}
and set $\Set{\Subst{z/a°c},\Subst{z/c°b}}$ is
represented by the formula $(z_a\AND \Not z_b\AND z_c) \OR (\Not z_a\AND z_b\AND
z_c)$.

Before we return to the application of BDDs, let us first consider the process of calculating the sequence $(\CalZ_i\mid i \geq
0)$. $\CalZ_0$ can be immediately derived from $\Phi_I(z)$. But how can we
compute $\CalZ_{i+1}$ given $\CalZ_i$ and $\CalF_{su}$\/? In order to answer this
question we define
\begin{equation}
  \label{e:deftta}
  \TT_{\phi(a)}(z,z') = [\Delta(z) \wedge z' \circ \vartheta^- = z
  \circ \vartheta^+] \enspace . 
\end{equation}
for each state update axiom $\Phi(a) \in \CalF_{su}$ of the form
\begin{displaymath}
\Delta(\STT(s)) \rightarrow \STT(\DO(a,s) \circ \vartheta^- = \STT(s)
\circ \vartheta^+
\end{displaymath}
Furthermore, for the set $\CalF_{su}$ we define
\begin{equation}\label{e:defT}
  \TT(z,z') = \bigvee_{\phi(a) \in \CalF_{su}} \TT_{\phi(a)}(z,z') \enspace .
\end{equation}
This definition is motivated by the following result, whose proof can again be
found in \cite{hoelldobler:stoerr:99a}

\begin{lemma} \label{l:tsua}
  Let $t$ and $t'$ be two constructor state terms and $\CalF \models \STT(s) = t$. Then,
  \begin{multline*}
  \CalF \models \STT(\DO(a,s))=t'\ \mbox{iff}\\
  \CalF_{un} \cup \CalF_{mset} \models \TT_{\phi(a)}(t,t')\
 \mbox{for some}\
  \phi(a) \in \CalF_{su}.
  \end{multline*}
\end{lemma}

Applying this lemma we are able to characterize our sequence $(\CalZ_i\mid i
\geq 0)$ without implicitly referring to $\CalF_{su}$ by the use of $\STT$:
\begin{equation}  \label{e:seqZZ} 
  \CalZ_n  =  \{ \sigma \mid \CalF \models \ZZ_n(z\sigma) \}, n \geq 0.
\end{equation}
where
\begin{align}
  \ZZ_0(z) &= \Phi_I(z)  \label{e:zz0}\\
  \ZZ_{i+1}(z) &= \Exists{z'}(\ZZ_i(z')\AND \TT(z',z)) \enspace . \label{e:zzstep}
\end{align}

The crucial point of our application of methods and techniques based on BDDs to
reasoning in the fluent calculus is the following: We could identify a class
$\FF$ of formulae over the alphabet underlying the fluent calculus and a
transformation $\CalB$ mapping each $F \in \FF$ to a propositional logic formula
$\CalB(F)$ such that (i) the class is expressive enough to represent interesting
entailment problems wrt the fluent calculus (namely, it contains the formulae
like $\ZZ_i$ defined in \eqref{e:zz0} and \eqref{e:zzstep}) and (ii) the
following result holds:

\begin{lemma} \label{l:btransl}
  Let $F \in \FF \cup \{\Phi_I(z),\ \Phi_G(z)\}$ and
  $\sigma$ a constructor state substitution such that $F\sigma$ does not
  contain any free variables. Then,
  \begin{displaymath}
    \CalF_{un} \cup \CalF_{mset} \models F \sigma\ \ \mbox{iff}\ \ 
    \CalB_S(\sigma) \models \CalB(F).    
  \end{displaymath}
\end{lemma}
The axiom set $\CalF_{un} \cup \CalF_{mset}$ contains the basic equational
theory behind the fluent calculus fragment used in this paper, and describes the
semantics of $\circ$. The precise definition of $\FF$ and $\CalB$ as well as the
proof of this lemma is beyond the scope of this paper and we refer the
interested reader to \cite{hoelldobler:stoerr:99a} or
\cite{hoelldobler:stoerr:00a} for the details. The class $\FF$ is subset of all
fluent calculus formulae consisting of restricted forms of equations of type
$\Sst=\Sst$ without use of function symbol $\STT$, as well as boolean
combinations of these and a restricted form of existential quantification over
\Sst\ variables.

Applying the translation $\CalB$ to the sequence of formulae $(\ZZ_i(z)\mid i
\geq 0)$ we obtain a procedure for calculating the
sequence $(\CalZ_i \mid i \geq 0)$ as follows. Let $\{f_1,\ldots,f_n\}$ be the
finite set of fluents in the alphabet underlying the fluent
calculus. Furthermore, let $F[z_1,\ldots,z_n]$ denote a propositional logic
formula $F$ built over the propositional variables $z_1,\ \ldots,\ z_n$. The
sequence $(Z_i \mid i \geq 0)$ of propositional logic formulae corresponding
to $(\CalZ_i \mid i \geq 0)$ is defined by
\begin{eqnarray}
  Z_0[\zv] & = & \CalB(\Phi_I(z)) \label{e:start} \\
  Z_{i+1}[\zpv] & = & \Exists{\zv}\ Z_i[\zv] \AND \CalB(\TT(z,z'))[\zv,\zpv],  \label{e:step}
\end{eqnarray}
where $\zv$ is the vector $z_{f_1},\dots,z_{f_n}$ of
propositional variables used to encode $z$ and $\Exists{\zv} F$ is an abbreviation for
$\Exists{z_1}\dots\Exists{z_n} F$ with
\begin{displaymath}
  \Exists{z_i} F = F\Subst{z_i/\False} \OR F\Subst{z_i/\True} \enspace .
\end{displaymath}

The propositional formulae $(Z_i \mid i \geq 0)$ are exactly the translations of
the fluent calculus formulae $(\ZZ_i \mid i \geq 0)$ by means of $\CalB$, and
thus, because of lemma~\ref{l:btransl}, representations of the sequence of sets
$(\CalZ_i \mid i \geq 0)$.

From~(\ref{e:start}) and~(\ref{e:step}) the so called {\em forward
pass \/} of our planning algorithm for computing the sequence $(\CalZ_i \mid i \geq 0)$ can be
derived:
\begin{enumerate}
\item Define $\CalZ_0$, in form of the BDD--representation of $Z_0$, such that
  it contains only the initial state of the reasoning problem.
\item Recursively calculate $\CalZ_{i+1}$, in form of the BDD-representation of
      $Z_{i+1}$, based on $Z_{i}$ and $\CalB(\TT(z,z'))$, until either $\CalZ_i$ overlaps with the set
      $G$ of goal states, in which case the reasoning problem is successfully solved or
      no new states are generated, in which case the reasoning problem is unsolvable.
\end{enumerate}
\begin{figure}[h]
  \begin{center}
    \hspace{-3mm}{\setlength{\unitlength}{4150sp}\renewcommand{\setlength}[2]{}
      \input{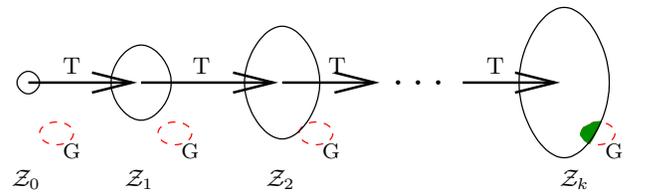}}
    \caption{\small The forward pass of our algorithm. After $k$
    steps the sets $\CalZ_k$ and $G$ overlap.}\label{f:forward}
  \end{center}
\end{figure}

The algorithm is illustrated in Fig.~\ref{f:forward}. Starting from the
initial state all reachable states are generated. The algorithm terminates as
soon as this set of states overlaps with the set of goal states or can no
longer be expanded. The inclusion of a special $\NOOP$ Action, which leaves the state unchanged, ensures that
the sets consecutively grow larger such that we know that all states have been visited iff
$\CalZ_i=\CalZ_{i+1}$. Alternatively to the inclusion of a $\NOOP$ Action, one can check for cycles in the sequence of sets and whether the sets
become empty.\footnote{When applying \emph{ frontier simplification}, as discussed later, the set $Z_i$
  becomes empty when all states have been considered.}
In either case the algorithm is guaranteed to terminate after at most $2^n-1$ steps, where $n$ denotes the
number of fluents, since the number of reachable states is at most $2^n$ and thus the length of the shortest
plan can be at most $2^n-1$ (such that every state is visited once).

If the forward pass terminates successfully, then in a second step a
shortest plan is constructed. This is done by choosing a state from $G \cap
\CalZ_k$ and searching for a chain of states through which this state can be
reached from the initial state. This is done by iterating backwards through
the sets $\CalZ_i$ generated by the forward pass algorithm. Because this second
step is a computationally (relatively) inexpensive part, we refer the
interested reader to \cite{hoelldobler:stoerr:99a}, where also the soundness and completeness of
the combined algorithm is established.

\section{Optimizations}\label{s:opt}

The planning approach described above is an implicit\footnote{It is called
implicit because the calculated sets of states are never explicitly enumerated, but
represented as a whole by a BDD, whose size depends more on the structure of the set, than on its actual
size.} breadth first search. In each single step 
we search the whole breadth of the search tree in depth~$i$. The sets $\CalZ_i$
can get quite complex and their BDDs quite large. Even more so, the size of
the BDD for $\CalB(\TT(z,z'))$, which describes the relation between the
$\CalZ_i$, can quickly become too large to be handled in a graceful manner. Our
approach shares this problem with related model checking algorithms.
Thus,
a number of techniques were invented to limit a potential explosion in its
size. In the sequel some of these techniques and their effects are discussed.

\subsection{Variable Order}
\label{s:order}

It is well known that the variable order used in a BDD has a large influence
on the size of the BDD.  Unfortunately it is still a difficult problem to find
even an near optimal variable order.\footnote{The problem to find the optimal
variable order is NP-complete.} Often, a good and acceptable variable order is
found by empiric knowledge and experimentation. A general rule is to group variables viz. fluents, which
directly influence each other, together. In particular, the variables $z_f$ and $z_f'$ occurring in
$\CalB(\TT(z,z'))$ should be ordered next to each other order. But
how should these variable groups be arranged? An ordering we call {\em sort ordering\/} led to good results in
several 
reasoning problems (see Tab.~\ref{t:sortordered}). The idea underlying the
sort ordering is to group fluents by their arguments. For example, in the
\textsc{Gripper} class the fluents $\AT(\BONE,\RA), \AT(\BONE,\RB),
\CARRY(\BONE,\GA), \CARRY(\BONE,\GB)$ should be grouped together, because they
share the argument $\BONE$. Remember that the fluent calculus is sorted.
The sort ordering works as follows. First one considers the argument of each fluent which belongs to the
largest sort and sorts the fluents according to this argument. 
The remaining ambiguities are resolved by considering the argument of the second largest sort and so
forth as well as the leading function symbol.
For some domains Tab.~\ref{t:sortordered} shows some almost dramatic
improvements in the size of the BDDs for sort ordering if compared to a simple
lexical ordering. The latter results in grouping fluents with the same
leading function symbol together. For some domains, however, there is little
or no improvement; this is usually the case when there are no large sets of
objects as parameters for fluents.

  \begin{table}[htb]
    \begin{center} \small
      \begin{tabular}{| l|r|r|r|r} \hline
        Problem & \textsc{Gripper} & {\sc Blocksworld} & {\sc get--paid}\\
        & \textsc{\sf\footnotesize (20 Balls)} & {\sf\footnotesize (8 Blocks)} & \\
        \hline
        lexical & 217409 & 206995 & 25633\\
        sort ordered & 3087 & 23373 & 38367 \\ \hline
      \end{tabular}
      \caption{\small The Size of the BDD for $\CalB(\TT(z,z'))$ with an ordering of the variables by
        name (lexical) or with the sort ordering  heuristic. The problems are from the planning problem
        repository \cite{pddlrepository}.}
      \label{t:sortordered}
    \end{center}
  \end{table}
\subsection{Partitioning of the Transition Relation}\label{s:partbdd}

The maximal size of a BDD is exponential in the number of propositional
variables it contains. Thus, the BDD representing $\CalB(\TT(z,z'))$, which
contains twice as many propositional variables as the BDDs representing the
$\CalZ_i$, is prone to get very large. A way to reduce this problem is to divide
the disjunction $\TT(z,z')$ into several parts $\TT_1,\dots,\TT_n$, which
correspond to subsets of the state update actions.

Let
$\CalF_{su,1},\ ,\dots,\ \CalF_{su,k}$ be a partition of $\CalF_{su}$
and define for all $1 \leq i \leq k$
\begin{gather*}
  \TT_i(z,z') = \bigvee_{\phi(a) \in \CalF_{su,i}} \TT_{\phi(a)}(z,z')
\end{gather*}
such that
\begin{math} \displaystyle
  \TT(z,z')= \bigvee_{i=1}^k \TT_i(z,z') \enspace .
\end{math}
Thus, \eqref{e:step} is modified to
\begin{multline}\label{e:steppart}
  Z_{i+1}[\zpv] = \\
  \bigvee_{i=1}^k \Exists{\zv} \bigl(Z_i[\zv] \AND \CalB(\TT_k(z,z'))[\zv,\zpv]\bigr) \enspace .
\end{multline}
\begin{figure}[h]
  \begin{center}
    {\setlength{\unitlength}{4150sp}\renewcommand{\setlength}[2]{}
      \input{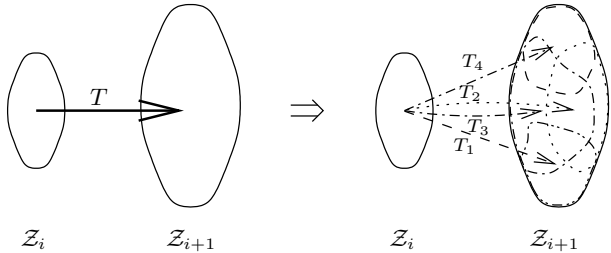} \vspace{-3mm}}
    \caption{ \small The partitioning of the transition relation. Each
    of the codomains of $T_1$,\ $T_2$,\ $T_3$ and $T_4$ is significantly smaller
    than the codomain of $T$.} \label{f:transpart}
  \end{center}
\end{figure}
Fig.~\ref{f:transpart} illustrates the partitioning of the transition relation.

The positive effect of the partitioning is that the actions in each subset effect
only a subset of all fluents.  Because the maximal size of a BDD is
exponential in the number of propositional variables, the sum of the sizes of
the BDDs corresponding to the partition may be significantly smaller than the
size of the original BDD.

\begin{figure}[p]
    \hspace{-0cm}\epsfig{file=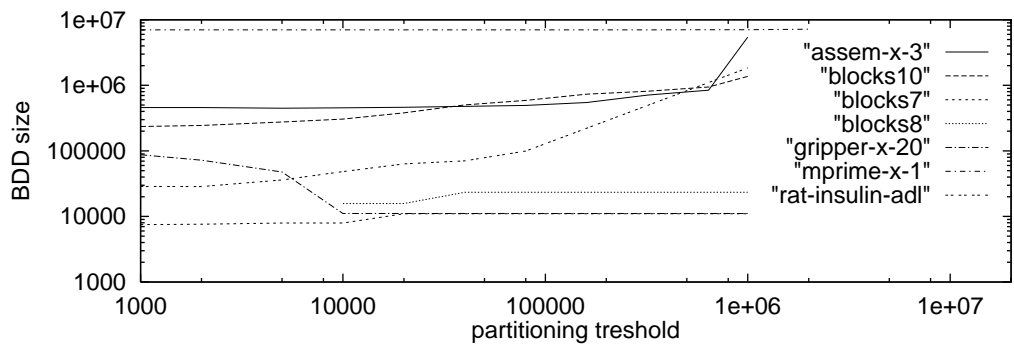}
    \caption{The sum of the sizes of the BDDs used to represent the transition relation in dependence
      on the parameter ``partitioning threshold''.}
    \label{f:splitbddsize}
\end{figure}

In our implementation the number of partitions is adaptive: first the BDDs $\CalB(\TT_{\phi(a)}(z,z'))$ for
every single action are constructed, then they are combined until a parameter ``partition treshold'' is
exceeded.  In the experiments, partitioning led to a reduction of needed memory in most of the tested problems
as shown in Fig.~\ref{f:splitbddsize} at the end of the paper.

\begin{figure}[p]
    \hspace{-0.2cm}\epsfig{file=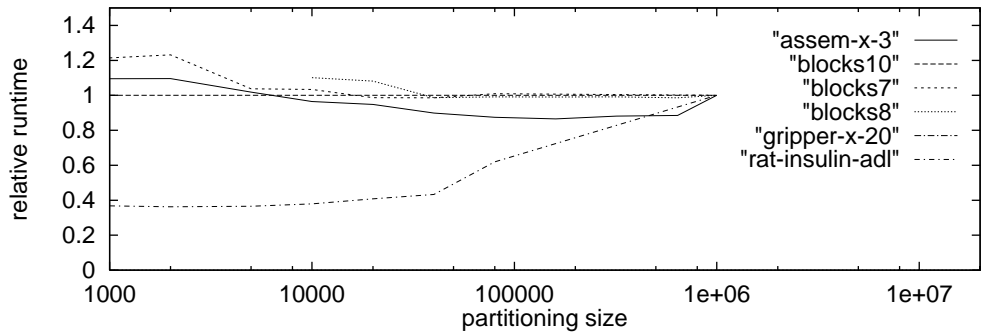}
    \caption{Effects of the parameter ``partitioning threshold'' on the
      calculation time for several problems. The time is relative to the time taken when no partitioning is
      done.}
    \label{f:split}
\end{figure}

On the other hand, a reduction in memory size does not necessarily lead to a
reduction in calculation time as the results depicted in Fig.~\ref{f:split}
indicate. According to equation \eqref{e:steppart} the various parts of the
partitioned transition relation have to be put together, and this takes time. 
Nevertheless splitting can be useful even if the computation time
increases, because of the reduction of the needed memory to store the
BDDs. For example in the case of {\sc mprime--x--1} the problem was not
manageable under our memory constraints without partitioning the transition
relation.

The idea to partition BDDs can also be applied to the BDDs representing $Z_i$.
We have not yet explored this idea, because in our test problems these BDDs
were only moderately large (i.e., up to 100.000 nodes).

We have also implemented an optimization technique called {\em frontier
  simplification\/} \cite{clarke:etal:94}. This technique explores the fact, that the
algorithm for solving the entailment problem in the fluent calculus works also
if the following two conditions are enforced for all $i \geq 0$: 
\begin{itemize}
\item The set $\CalZ_i$ contains all states which may be reached by executing $i$
      actions, but not by executing less than $i$ actions. 
\item the set $\CalZ_i$ does not contain any states which cannot be reached by
  executing at most $i$ actions.
\end{itemize}
The sets $\CalZ_i$ can be chosen freely within these limitations. Hence, it is
desirable that the algorithm chooses the $\CalZ_i$ such that their BDD
representations are as small as possible. In our experiments frontier simplification lead to moderate
improvements (i.e. up to 40\
improved computation times, because the time saved for the computation of the recursion equation
\eqref{e:step} was outweighted by the additional effort spent for the reduction of the BDDs.

\section{Results on the \textsc{Gripper} Class}  \label{s:gripper}

\begin{figure}[p]
  \hspace{-0.1cm}\epsfig{file=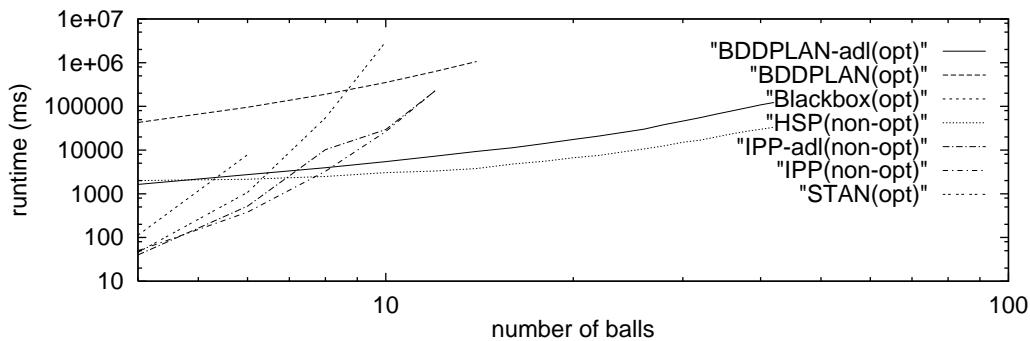}
  \caption{Runtimes of different planners on the \texttt{Gripper} problem (in milliseconds) with different
    numbers of balls. Planners marked with \texttt{opt} provided optimal (i.e. shortest) plans, planners
    marked with \texttt{-adl} work on the sorted version of the domains, the others on the STRIPS-version.}
  \label{f:compare}
\end{figure}

The problems of the \textsc{Gripper} class were quite hard problems for the
planners taking part in the AIPS98 competition. Their difficulty is rooted in
the combinatorial explosion of alternatives due to the existence of two
grippers.  In Fig.~\ref{f:compare} the runtimes of these planners\footnote{See
  \textsf{http://ftp.cs.yale.edu/pub/mcdermott/aipscomp-\\ results.html}.} are compared
to our system, \BDDPLAN.\footnote{The runtime of \BDDPLAN\ is measured on a
different machine, so the comparison is only accurate up to a constant factor.} Only one
planner (HSP) was able to solve all of the problems of this class, but it
generated only suboptimal plans by using only one of the two grippers, whereas
\BDDPLAN\ generates the shortest possible plan by design.

\section{Discussion}

We have presented in this paper our preliminary findings in applying BDD
techniques as an implementation tool for reasoning about situations, actions
and causality in the fluent calculus, and discussed several techniques that
have been successfully used to improve the performance of the implementation.

We tested our implementation using the problems of the planning contest on
AIPS98 and have received mixed results so far. As discussed in
section~\ref{s:gripper}, our planner performed very good in the {\sc Gripper}
class: It was able to provide the shortest solutions to even the most
difficult problems posed in this class, whereas the planners which have
participated in the competition were only able to solve but the simplest
problems or, in the exceptional case of HSP, provided sub--optimal solutions
ignoring the second gripper of the robot. In some other problem classes, however,
our implementation did not outperform existing systems. On the other hand, we
have just started to investigate optimization techniques and will continue to
do so in the future.

At present, our algorithm (described in more detail in
\cite{hoelldobler:stoerr:00a,hoelldobler:stoerr:99a}) is closely related to
model checking algorithms \cite{burch:etal:92} which perform symbolic breadth
first search in the statespace. It generates a series $(Z_i \mid i \geq 0)$ of
propositional formulae represented as BDDs, which encode the set of answer
substitutions $\sigma=\{z/t\}$ for the fluent calculus formulae
$\CalF\models(\exists (a_i)_{1 \leq i \leq n})\ z\sigma = \STT(a_n \ldots a_1
S_0)$, which represents sets of states $t$ reachable after the execution
of a sequence of actions $(a_i)_{1 \leq i \leq n}$ of length $n$, until there is
a goal state among the states encoded. The formulae $Z_i$ are generated
recursively by applying the propositional encoding of a transition relation
$\TT(z,z')$.
  
The optimization techniques presented in this paper do not change the
principle of breadth first search the algorithm is based on. This has the
pleasant effect that
\begin{itemize}
\item the algorithm is complete in the sense that it always either finds the
  shortest plan or is able to prove that there is no plan, and
\item it is possible to reuse the results of the computationally
      intensive forward pass stage, in which the sequence of sets of reachable
      states $(\CalZ_i \mid i \geq 0)$ is constructed, to either create
      many solutions to the same reasoning problem or to solve
      multiple reasoning problems with the same initial state.
\end{itemize}

On the other hand, in order to speed up the search it seems one should give up
the concept of breadth first search and explore {\em interesting\/} parts of the
search space first. This can be done without giving up completeness by stepwise
adding actions to the transition relation, which seem heuristically relevant for
reaching the goal, and explore the subtrees of the search space generated by
these actions first. This concept is similar to abstraction in planning
\cite{knoblock:94} and is topic of future research.

It should be noted that although we have presented our algorithm in such a way
that there is only a single initial state (i.e., the set $\CalZ_0$ is unitary),
the algorithm itself is not restricted to this case. If the initial situation is
only incompletely specified then there are several initial states, which leads
to a set $\CalZ_0$ containing more than one element. However, a straightforward
application of the algorithm to such a non--unitary set $\CalZ_0$ would result
in a ``brave'' reasoning process in the sense that the plan generated works for
at least one of the initial states, but is not guaranteed to work for the
others.

There is a number of approaches in planning that are based on propositional
logic \cite{weld:99}. Many of the most successful are rooted in the planning as
satisfiability \cite{kautz:selman:96} and Graphplan \cite{blum:furst:97} or
both. Our algorithm is similar to Graphplan in that it builds up a data
structure for each level, which describes the states reachable after the
execution of $n$ actions, (though Graphplan admits the parallel execution of
multiple actions in a time step if they do not interfere.)  Unlike Graphplan,
that gives only an upper bound of the the set of states reachable by its mutex
mechanism, our algorithm computes an exact symbolic representation of this set.
Consequently, the plan extraction process is deterministic and no backtracking
is needed.

In contrast to algorithms based on planning as satisfiability (SATPLAN) and
Graphplan the algorithm presented here is not limited to the generation of
polynomial length plans and is complete. On the other hand, each time step may
take space exponential space, since the maximum size of BDDs is $O(2^n)$ for $n$
propositional variables. However, the experimental results achieved so far indicate
that in practice the BDDs are much smaller than the theoretical limit.

Still, the size of the encountered BDDs is the main problem limiting the
scalability of the algorithm and is an topic of further research. Since the
maximum size of BDDs is exponential in the number of propositional variables,
the reduction of this number is a foremost concern. By design our algorithm
avoids the unfolding of all time steps of the plan into disjunct sets of
propositional variables, as in the case of SATPLAN and Graphplan, since all
time steps are treated separately. Moreover we are able to omit the variables
encoding actions easily, since we are not restricted to a clausal form of the
formulas we are working with, and the actions can be reconstructed from the
sequence of states. The encoding we use at present is ``naive'' in the sense
that each fluent corresponds to a single propositional variable. We assume that
the use of domain dependent properties of fluents provides a large space for
improvements, as discussed in \cite{edelkamp:helmert:99} for the BDD based
planning system \emph{Mips}, which is used to explore automated generation of
efficient state encodings for STRIPS/ADL/PDDL planning problems and the
implementation of heuristic search algorithms with BDDs.

Depending on the task, it seems to be inevitable to encode the actions in the
case of non--deterministic domains, as in the work of \cite{cimatti:etal:98}.
Their system generates so--called {\em universal\/} plans, which consist of a
state--action table that contains for each state the action, which leads to
the goal in the shortest way. This approach opens new possibilities in
generation of plans for non--deterministic domains. However, considering the
case of deterministic domains, we conjecture, that this approach is limited to
less complex reasoning problems in comparison to state--only encodings,
because, additional to the states before and after the execution, the executed
actions have to be encoded into the transition relation as well. This leads to
a considerable increase in the number of propositional variables and,
consequently, in the maximal size of the BDDs.  But we have not yet performed
direct comparisons to bolster this conjecture.

The translation used in our approach to map fluent calculus entailment problems
to propositional logic is tailored to a specific class of fluent calculus formulae,
which is just large enough to specify the considered class of planning
problems. However, it seems likely that there is a more general way to
translate the formulas of a larger fragment of the fluent calculus while
keeping the restriction to propositional fluents, such that we could introduce
recent work on the fluent calculus like ramification
\cite{thielscher:98,thielscher:aij98} into our planner without modifying the
translation and the proofs. The concept of ramification within the fluent
calculus involves a limited use of constructs of second order logic, namely a
calculation of the transitive closure of a relation over states, but this does
not seem to pose a difficult problem as the set of states is finite and there
are algorithms to compute this transitive closure using BDDs
\cite{clarke:etal:94}.

To sum up, our BDD based implementation shows some promising initial results
but it is too early to completely evaluate it yet.

\bibliography{/home/stoerr/lit/hapslit,/home/stoerr/lit/mylit,/home/sh/Lit/abbrev,/home/sh/Lit/ag,/home/sh/Lit/hm,/home/sh/Lit/nz,/home/sh/Lit/h,lit}

\onecolumn
\twocolumn

\end{document}